\documentclass{article}

\usepackage{PRIMEarxiv}

\usepackage[utf8]{inputenc} 
\usepackage[T1]{fontenc}    
\usepackage{hyperref}       
\usepackage{url}            
\usepackage{booktabs}       
\usepackage{amsfonts}       
\usepackage{nicefrac}       
\usepackage{microtype}      
\usepackage{lipsum}
\usepackage{fancyhdr}       
\usepackage{graphicx}       
\graphicspath{{media/}}     

\usepackage{times}
\usepackage{multicol}
\usepackage{subcaption}
\usepackage[labelfont=bf]{caption}
\usepackage{textcomp}
\usepackage{arydshln}
\usepackage{gensymb}
\usepackage[table]{xcolor}
\usepackage{graphicx}
\usepackage{multirow}
\usepackage{booktabs}
\usepackage{url}
\usepackage{paralist}
\usepackage{amssymb}
\usepackage{lipsum}  
\usepackage{collcell}
\usepackage{xcolor}
\usepackage{etoolbox}

\usepackage[T1]{fontenc}
\usepackage[utf8]{inputenc}
\usepackage[american]{babel}
\usepackage[table]{xcolor}
\usepackage{collcell}
\usepackage{hhline}
\usepackage{pgf}
\usepackage{multirow}
\usepackage{color, colortbl}
\usepackage{adjustbox}
\usepackage{esvect}
\usepackage{tikz}
\usepackage[autostyle]{csquotes}
\let\labelindent\relax
\usepackage[]{enumitem}

\def\colorModel{hsb} 

\newcommand\ColCell[1]{
  \pgfmathparse{#1<50?1:0}  
    \ifnum\pgfmathresult=0\relax\color{white}\fi
  \pgfmathsetmacro\compA{.52}   
\pgfmathsetmacro\compB{#1/40} 
\pgfmathsetmacro\compC{1-#1/500}        
  \edef\x{\noexpand\centering\noexpand\cellcolor[\colorModel]{\compA,\compB,\compC}}\x #1
  } 
  
\newcolumntype{E}{>{\collectcell\ColCell}m{0.3cm}<{\endcollectcell}}  

\newcommand{\ie}{\textit{i.e.}, }
\newcommand{\eg}{\textit{e.g.}, }
\newcommand{\etal}{\textit{et al.}, }

\usepackage{csquotes}

\makeatletter
\DeclareRobustCommand{\change}{%
  \@bsphack
  \leavevmode
  \color{red}%
  \@esphack
}
\DeclareRobustCommand{\stopchange}{%
  \@bsphack
  \normalcolor
  \@esphack
}
\makeatother

\title{Diver Identification Using Anthropometric Data Ratios for Underwater Multi-Human-Robot Collaboration
\thanks{This work was supported by the NSF under Grant IIS-2220956. $^*$The authors make equal contributions to the work.}}

\author{
  Jungseok Hong$^{*}$, Sadman Sakib Enan$^{*}$, Junaed Sattar \\
  Department of Computer Science \& Engineering and the Minnesota Robotics Institute \\
 University of Minnesota \\
  Minneapolis, MN USA\\
  \texttt{\{jungseok, enan0001, junaed\}@umn.edu} \\
}

\begin{document}
\maketitle
\begin{abstract}
Recent advances in efficient design, perception algorithms, and computing hardware have made it possible to create improved human-robot interaction (HRI) capabilities for autonomous underwater vehicles (AUVs). To conduct secure missions as underwater human-robot teams, AUVs require the ability to accurately identify divers. However, this remains an open problem due to divers' challenging visual features, mainly caused by similar-looking scuba gear. In this paper, we present a novel algorithm that can perform diver identification using either pre-trained models or models trained during deployment. We exploit \textit{anthropometric} data obtained from diver pose estimates to generate robust features that are invariant to changes in distance and photometric conditions. We also propose an embedding network that maximizes inter-class distances in the feature space and minimizes those for the intra-class features, which significantly improves classification performance. Furthermore, we present an end-to-end diver identification framework that operates on an AUV and evaluate the accuracy of the proposed algorithm. Quantitative results in controlled-water experiments show that our algorithm achieves a high level of accuracy in diver identification. 
\end{abstract}



\section{Introduction}
\begin{figure}  
    \vspace{2mm}
    \centering
    \includegraphics[width=0.6\linewidth]{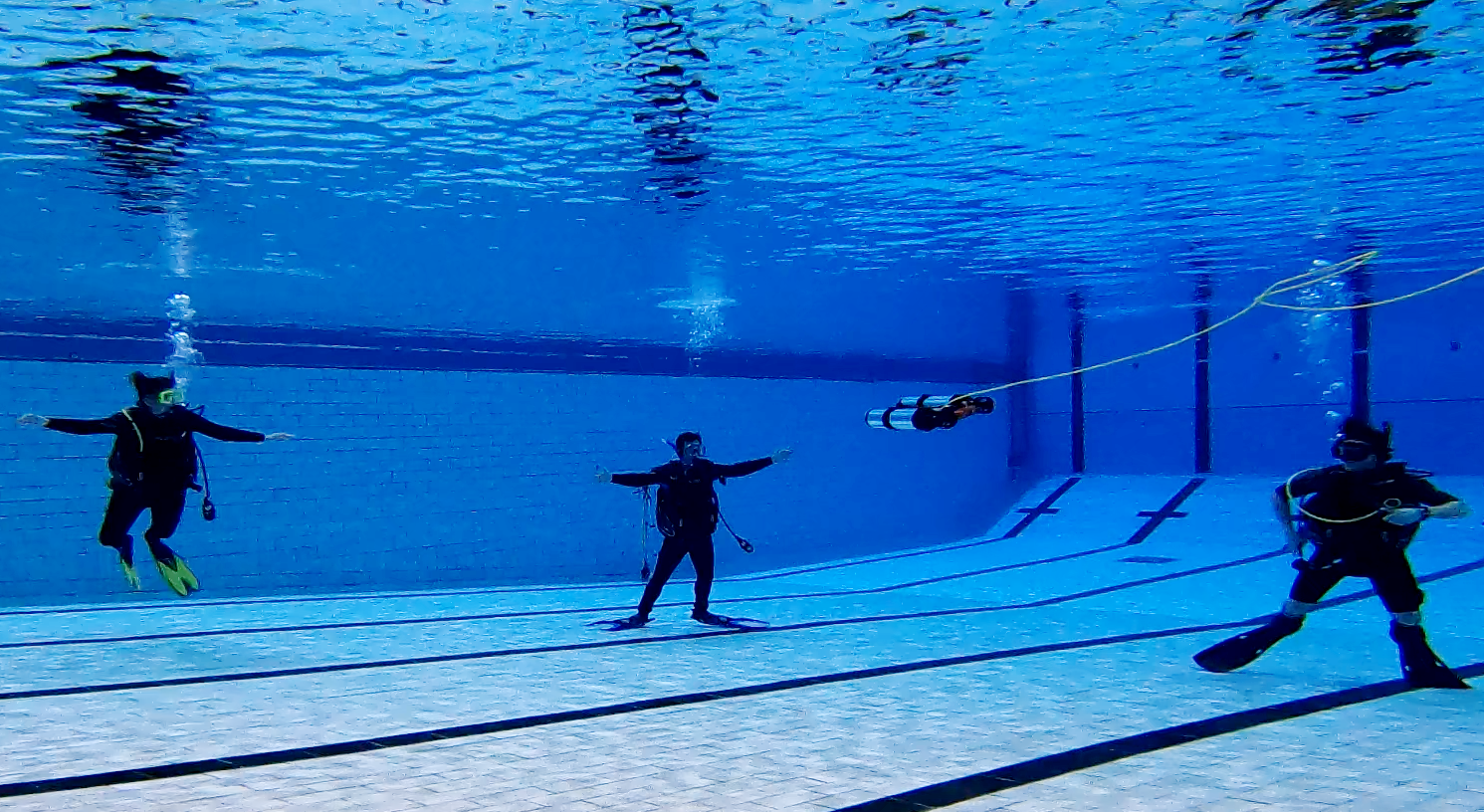}
    \caption{An example scenario of a robot identifying divers with our proposed framework by using the \textit{anthropometric} features extracted from the divers.}  
    \label{fig:intro}
    \vspace{-4mm}
\end{figure}

In recent years, underwater robotics has advanced significantly, mainly driven by increased onboard computational power, improved perception, and enhanced HRI capabilities of AUVs. 
Such advances allow AUVs to be used in both standalone (\eg environmental monitoring~\cite{bayat2017environmental}) and collaborative missions (\eg diver following~\cite{islam2019diverfollowing}). Underwater HRI has the potential to be extended for more advanced human-robot collaborative missions (\eg shipwreck and pipeline inspection~\cite{bingham2010robotic}) where robots can rely on human expertise while reducing risks posed to humans. However, there are still challenges that need to be addressed to use HRI capabilities for such missions, with diver identification specifically posing as one of these obstacles. In general, AUVs have limited operating time underwater due to constraints caused by environmental (\eg rapid current and water turbidity) and operational factors (\eg limited battery time). Taking conflicting instructions from divers can delay and jeopardize missions. For instance, in a multi-diver single-robot collaboration scenario (as shown in Fig.~\ref{fig:intro}), the ability to correctly identify divers would enable the robot to manage individual diver data and instructions separately. This would result in enhanced operational efficiency.

To address the diver identification problem, researchers have proposed various solutions, including fiducial markers~\cite{sattar2007fourier}, unique hand signs~\cite{islam2018dynamic}, and acoustic communication devices~\cite{mivskovic2016human}. 
However, with these methods, the diver must carry additional items or memorize unique hand signals, which can add to the already-elevated cognitive and physical burdens. Losing or forgetting such means of identification compromises the entire mission. To address these issues, a recent work uses facial cues to identify different divers by synthetically removing the dive mask and breathing apparatus from their face to compute unique facial features~\cite{hong2020visual}. However, the proposed method requires the collection of diver face data prior to the mission. Also, it can fail due to facial distortion at the edges of the masks and the area surrounding the lips caused by wearing the mask and breathing apparatus. 

In this work, we propose a framework that uses distance and photometry-invariant \textit{anthropometric} features for diver identification and demonstrate that it works robustly underwater in data-scarce conditions. 
Our method can be deployed with either pre-trained models (\textbf{offline framework}) or models trained during deployment (\textbf{online framework}). This allows the AUV to identify divers even if their data is not present in our dataset. 
This enables efficient human-robot collaborative missions with robust diver identification by addressing the underwater domain-specific challenges and constraints introduced above.
\textbf{Anthropometric data (AD)} refers to the study of human body measurements, such as the width of shoulders, lengths of the lower and upper arms, and measurement of other limbs or limb sections. 
AD can be obtained even when divers' bodies and faces are heavily obscured. However, the AD values may change during the identification process, since the distance between the robot and the diver can vary. 
Therefore, we propose to use \textbf{anthropometric data ratios (ADRs)} instead, to represent divers uniquely, exploiting the property that ADRs will be invariant to the robot-diver distance. 

In our framework, the robot computes the ADR features by estimating diver poses and subsequent pose filtering.
Through experiments, we demonstrate that these ADR features can be projected into a $16$-dimensional ($16$-d) hyperspace where they show better separation. We achieve this by using an embedding network that maximizes inter-class feature distance and minimizes intra-class feature distance. We demonstrate the efficacy of the proposed method in a closed-water environment onboard a
physical AUV. We make the following contributions in this paper:

\begin{enumerate}
    \item We propose distance and photometric invariant features, called \textbf{anthropometric data ratios (ADRs)}, to create unique representations of scuba divers. 
    \item We introduce an embedding network that can project the ADR features to a $16$-d hyperspace where they form highly separable inter-class clusters, resulting in superior classification performance.
    \item We use anthropometric data statistics to filter out erroneous diver pose estimations. 
    \item We propose a diver identification algorithm for underwater robots to use the collected features to train a number of models both offline and online.
    \item We measure the efficacy of the system on a physical AUV in a closed-water environment.
\end{enumerate}
\section{Related Work}
One of the preliminary works on person identification was described in~\cite{bertillon1896signaletic} where the proposed method used AD values to identify different individuals. As the technology evolved, AD-based identification methods were replaced by systems that use biometric information, such as fingerprints, voice, and iris scans~\cite{drozdowski2019computational}. Spectral information extracted from electroencephalogram (EEG) signals can also be used to identify a person~\cite{wilaiprasitporn2020affective}. Furthermore, person identification using facial cues has seen incredible success~\cite{kortli2020face,deng2019arcface}, especially with the availability of large-scale human face datasets~\cite{yang2016wider, guo2016ms, bae2023digiface}, superior computing power, and advances in deep convolutional neural networks~\cite{bengio2017deep}. Alternatively, gait detection can also be used to uniquely identify a person, as it has been found that walking patterns have a high correlation to a subject's identity~\cite{khan2021gait}. 
\begin{figure*}  
    \vspace{2mm}
    \centering
    \includegraphics[width=\linewidth]{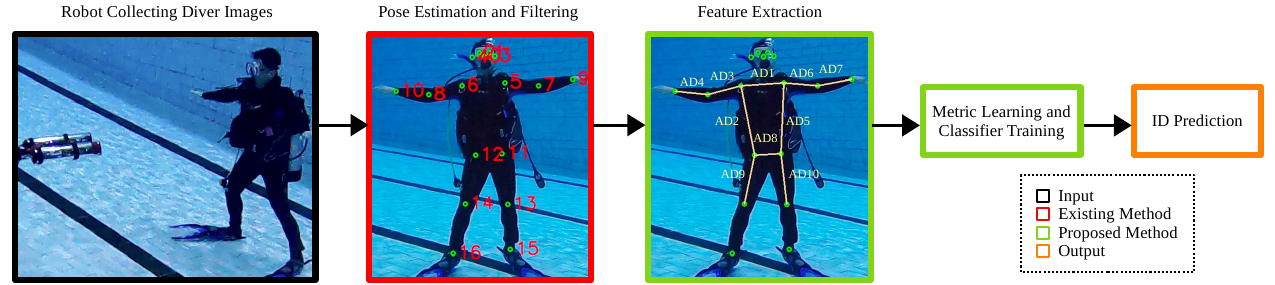}
    \caption{Given an RGB image as input, our proposed algorithm first extracts features from the diver's pose. It then predicts the diver's ID using these highly distinguishable features. The numbers in the second sub-figure represent $17$ body keypoints. Additionally, the AD$\{1,\cdots,10\}$ in the third sub-figure are the extracted features.}
    \label{fig:id}
    \vspace{-4mm}
\end{figure*}


The above-mentioned techniques were developed for terrestrial applications. Person identification techniques in an underwater environment (\ie diver identification), on the other hand, are significantly less studied. In a recent work~\cite{xia2019visual}, the authors present a diver identification framework by leveraging features from spatial and frequency domains to uniquely represent different divers. In a similar line of work~\cite{de2020realtime}, the authors use appearance and location metrics to track divers for re-identification. However, in both of these papers, the divers need to be in motion to extract feature representations. In contrast, the work in~\cite{hong2020visual} proposes to use facial cues to identify divers for HRI, but achieves sub-par results because of the difficulty in extracting good facial features from heavily obscured diver faces.    

Recently, Munsell \etal~\cite{munsell2012person} show that AD can be used to identify different individuals. They demonstrate that these data are less sensitive to photometric differences and more robust to obstructions, \eg glasses, and hats. Similarly, the work in~\cite{andersson2015person} shows that anthropometric and gait features can be used to uniquely identify different people. 
In this paper, we also choose to use AD to represent and identify different scuba divers. To compute them, we rely on 2D pose estimation methods which can find the location of different human body joints. There are a number of 2D pose estimation approaches (\eg OpenPose~\cite{openpose}, trt\_pose~\cite{trtpose}, MediaPipe~\cite{lugaresi2019mediapipe}, DeepLabCut~\cite{mathis2018deeplabcut} to name a few) that achieve high accuracy in terrestrial applications. For our purpose, these methods generated a high number of incorrect pose estimations for the divers we used in our study, possibly due to water turbidity and challenging lighting conditions. 
A recent work~\cite{wang2020deep} uses high-resolution representation learning, by employing several high-to-low resolution convolutional branches in parallel, to achieve semantically rich and spatially precise pose estimations. We exploit this method in our work to estimate diver poses.
\section{Methodology}\label{sec:method}
\subsection{Diver Identification Algorithm}
Our proposed diver identification framework consists of four major components: (1) Pose Estimation and Filtering, (2) Feature Extraction, (3) Metric Learning, and (4) Classifier Training. We perform pose estimation on the image frames containing a diver and filter them using anthropometric data statistics. Then, we compute features from the filtered pose estimations to train different models to identify divers. Fig.~\ref{fig:id} visualizes this process.
    \begin{figure}
    \vspace{2mm}
        \centering
        \begin{subfigure}[b]{0.45\linewidth}   
            \centering 
            \includegraphics[width=0.9\linewidth
            ,trim={0 .8cm 0 1.8cm},clip]{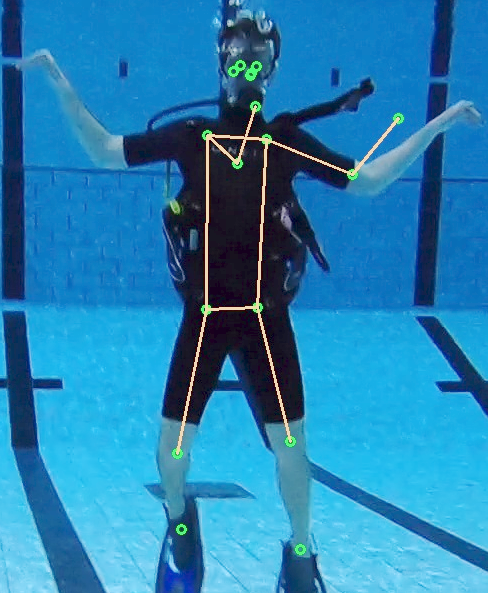}
            \label{fig:fail1}
        \end{subfigure}
        \begin{subfigure}[b]{0.46\linewidth}   
            \centering 
            \includegraphics[width=0.9\linewidth,trim={0 .8cm 0 1.8cm},clip]{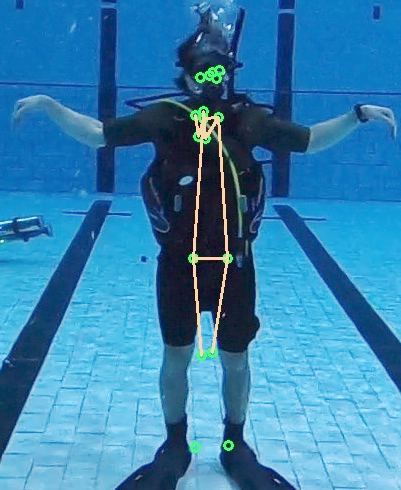}
            \label{fig:fail2}
        \end{subfigure}%
        \caption{Examples of pose estimation failures caused by rapid movements of divers' arms underwater. As can be seen, the predictions for wrists, elbows, and shoulder joints locations are inaccurate.}
        \label{fig:failure}
        \vspace{-4mm}
    \end{figure}

\subsubsection{Pose Estimation and Filtering}
To maximize the probability of getting stable pose estimation, our system requires the divers to take a standing-upright and frontal-facing position from the robot's perspective. We use HRNet~\cite{wang2020deep}, a robust 2D pose estimation method, to predict the joint locations of divers. While such posture requirements can help to produce good predictions, they cannot completely prevent the pose estimation method from making false predictions. Especially, rapid movements of scuba divers' arms often make it extremely difficult for the pose estimation algorithm to accurately predict the poses (see Fig.~\ref{fig:failure}). 
Originally HRNet estimates $17$ body keypoints, from which we select the following $10$ keypoints: \textit{left shoulder, right shoulder, left elbow, right elbow, left wrist, right wrist, left hip, right hip, left knee, right knee}. While including the remaining seven keypoints (\textit{i.e., nose, left eye, right eye, left ear, right ear, left ankle, right ankle}) could make our feature set richer, we observed that their predictions tend to be inconsistent across frames compared to the selected $10$ keypoints. Additionally, the unselected seven keypoints require the underwater robot to constantly capture the whole body of the diver, which may not be realistic during underwater missions due to environmental factors (\eg waves). 

To create robust ADR features from the predicted pose, erroneous data must be removed from the dataset. We achieve this by formulating a number of filtering conditions motivated by anthropometric data statistics~\cite{gordon1989anthropometric}. Specifically, we use the following filtering conditions, where $hw_{min}$, $k_{th}$, and $sw_{min}$ are hyperparameters.
\begin{itemize}
    \item hips are located lower than shoulders.
    \item knees are located lower than hips.
    \item hip width must be larger than $hw_{min}$.
    \item both hip to knee distances must be similar.
    \item both elbow to wrist distances must be similar.
    \item upper arm is slightly longer than the lower arm.
    \item distance between knees must be larger than $k_{th}$.
    \item shoulder-to-hip distance must be slightly longer than shoulder width.
    \item shoulder-to-hip distance cannot be twice as large as the hip-to-knee distance.
    \item shoulder width must be larger than $sw_{min}$.
    \item shoulder-to-hip distance must be larger than the sizes of the lower and upper arm.
\end{itemize}

We only keep the pose estimations that meet all the filtering conditions.

\begin{table}[t!]
    \vspace{2mm}
        \centering
        \caption{Our proposed embedding network, which takes $45$-d ADR features and project them into $16$-d features.} 
\begin{tabular}{p{0.15\linewidth}p{0.55\linewidth}}        
\toprule
Input:
 & ADRs ($F \times 45$)\\
\midrule
Layer 1
 & Linear $(45, 1024)$; Leaky ReLU; BN \\
\midrule
Layer 2
 & Linear $(1024, 512)$; Leaky ReLU; BN \\
\midrule
Layer 3
 & Linear $(512, 256)$; Leaky ReLU; BN \\
\midrule
Layer 4
 & Linear $(256, 16)$ \\
 \bottomrule \\
  \multicolumn{2}{l}{Here, $F$ = no. of filtered pose estimations.} 
\end{tabular}
    \label{tab:nn}
    \vspace{-4mm}
        \end{table}

\subsubsection{Feature Extraction} 
Once we obtain the filtered pose estimations, we calculate $10$ AD values using the $10$ body keypoints, as shown in the third sub-figure of Fig.~\ref{fig:id}. 
While the keypoint predictions are relatively stable, their locations in each frame constantly change as the robot moves. Therefore, keypoints and also the AD values are distance-variant. To address this issue and make the features distance-invariant, we propose to use ADRs as features to represent the divers. We compute the ADRs by taking the ratios of two different AD values, generating a total of $45$ ratios from unique pairs, excluding the pairs formed by identical ADs. We observed that the ADRs stay relatively consistent for each diver even when the distance between the robot and the diver changes. 
\begin{figure*}[ht!]
        \vspace{2mm}
        \centering
        \begin{subfigure}[b]{0.48\linewidth}   
            \centering 
            \includegraphics[width=\linewidth]{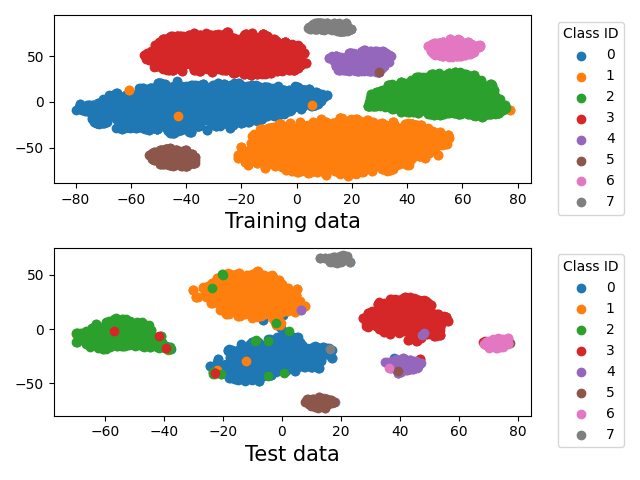}
            \caption{Trained on the All-class dataset.}
            \label{fig:all_cluster}
        \end{subfigure}
        \begin{subfigure}[b]{0.48\linewidth}   
            \centering 
           \includegraphics[width=\linewidth]{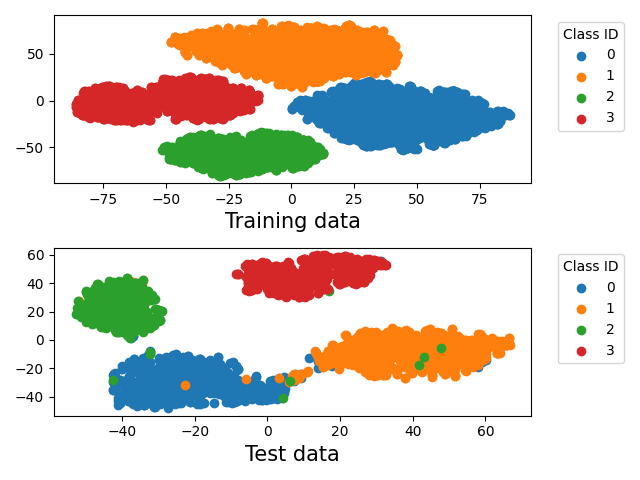}
            \caption{Trained on the Diver dataset.}%
            \label{fig:diver_cluser}
        \end{subfigure}%
        \caption{Results of the metric learning technique: clear separation among classes are seen. The $16$-d features are visualized in 2D plots using t-SNE technique. The top and bottom rows show the clustering results on the training and test data, respectively.}
        \label{fig:cluster}
            \vspace{-4mm}
\end{figure*}
\begin{table}[t!]
        \centering
        \caption{Network structure and online training capabilities of different models used for our diver identification framework.}
\begin{tabular}{lcccc}
    
\toprule

 \textbf{Model Name}* & \textbf{Embedding} & \textbf{Classification} & \textbf{Offline} & \textbf{Online}\\
\midrule
All\_KNN 
& - & KNN & \checkmark & -\\
\midrule
Diver\_KNN
& - & KNN & \checkmark & \checkmark\\
\midrule
All\_SVM
& - & SVM & \checkmark & -\\
\midrule
Diver\_SVM
& - & SVM & \checkmark & \checkmark\\
\midrule
All\_NN\_KNN
& NN & KNN & \checkmark & \checkmark\\
\midrule
Diver\_NN\_KNN
& NN & KNN & \checkmark & \checkmark\\
\midrule
All\_NN\_SVM
& NN & SVM & \checkmark & \checkmark\\
\midrule
Diver\_NN\_SVM
& NN & SVM & \checkmark & \checkmark\\
\midrule
All\_NN
& NN & NN & \checkmark & -\\
\midrule
Diver\_NN
& NN & NN & \checkmark & -\\
 \bottomrule \\
 
 \multicolumn{5}{l}{*All\_KNN and Diver\_KNN refer to the same model for online training.} 
 \\
 \multicolumn{5}{l}{Same goes for (All\_SVM, Diver\_SVM) and (All\_NN, Diver\_NN).} 
\end{tabular}
    \label{tab:all_models}
    \vspace{-4mm}
        \end{table}

\subsubsection{Metric Learning}\label{sec:metric_learning}
Metric learning~\cite{yang2006distance} is a method to learn a distance measure for maximally separating inter-class distances in the feature space, using an appropriate distance metric.
Examples of classical approaches are the K-nearest neighbor (KNN)~\cite{peterson2009k} and support vector machine (SVM)~\cite{noble2006support}. 
We apply metric learning when we train our proposed embedding network (as shown in Table~\ref{tab:nn}). We select the triplet loss~\cite{schroff2015facenet} as our loss function due to its simplicity and effectiveness. The loss function is formulated as follows:
\begin{equation}
    \mathcal{L}(A,P,N) = max (0, \mathcal{D}(A,P)-\mathcal{D}(A,N)+m)
    \label{eq:triplet}
\end{equation}
where $A$ is an anchor point (reference data), $P$ is a positive point, $N$ is a negative point, and $m$ is a predefined margin. 
By training our model with this loss function, we enforce intra-class data points (\ie $A$ and $P$) to stay close while increasing the distance between inter-class data points (\ie $A$ and $N$). In~\cite{schroff2015facenet}, Euclidean distance was used for the distance function $\mathcal{D}$. While this is one of the common distance metrics, it cannot capture nonlinearity in the data. Thus, we use cosine similarity as a distance metric in our work. 

Additionally, we created the following datasets for training the embedding network: (1) All-class dataset (four divers and four swimmers) and (2) Diver dataset (four divers only). The datasets are described in more detail in Sec.~\ref{sec:exp}. Our embedding network takes $45$-d features and outputs $16$-d feature vectors. Interestingly, upon plotting these $16$-d features from both training and test sets, they are observed to form well-defined clusters (see Fig.~\ref{fig:cluster}). We use the t-SNE~\cite{van2008visualizing} technique to plot the $16$-d features in a 2D plot.

\begin{figure*}  
    \vspace{2mm}
    \centering
    \scalebox{0.75}{\input{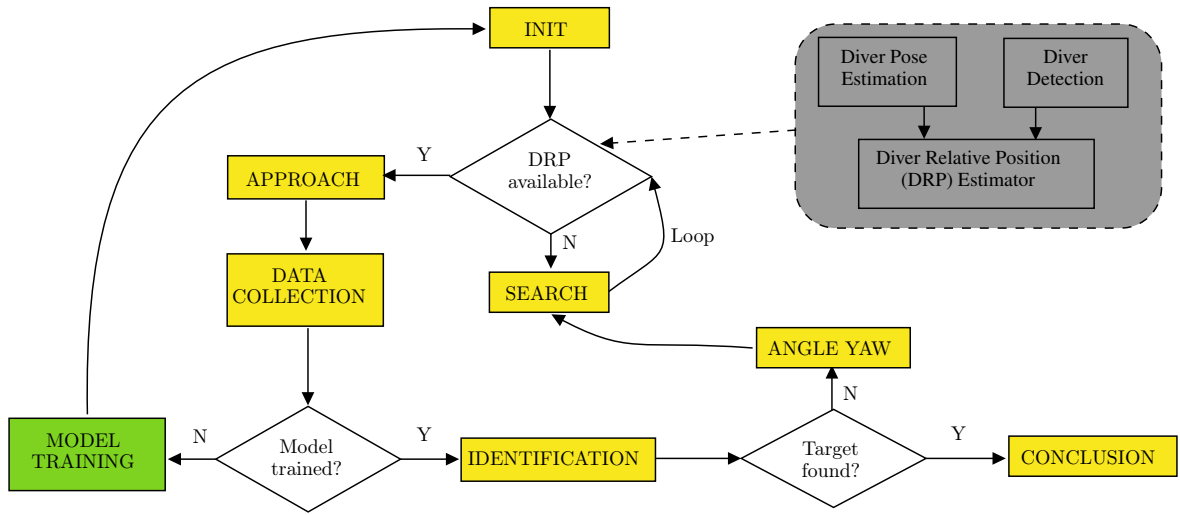}}
    \caption{Our proposed diver identification framework. The states in yellow boxes are common between the offline and online frameworks. Whereas, the state in the green box is used only in the online framework to perform model training during mission. The gray colored box shows the DRP estimator which runs continuously in the background.}  
    \label{fig:algoframe}
    \vspace{-2mm}
\end{figure*}

\subsubsection{Classifier Training}
We use $10$ models to evaluate the performance of the diver identification algorithm (as shown in Table~\ref{tab:all_models}). The model names starting with \textit{All} are trained with the All-class dataset, and those starting with \textit{Diver} are trained with the Diver dataset. For All\_KNN, Diver\_KNN, All\_SVM, and Diver\_SVM models, we use only either KNN or SVM to train for classification without using an embedding network. For All\_NN\_KNN, Diver\_NN\_KNN, All\_NN\_SVM, Diver\_NN\_SVM, All\_NN, and Diver\_NN models, we first pass the raw $45$-d ADR features through our embedding network to generate highly distinguishable $16$-d features. We then train classifiers with the $16$-d features. In the case of the All\_NN and Diver\_NN models, we use a two-layer neural network, attached with a final softmax layer to perform the multi-class classification using the $16$-d features.

\subsection{Diver Identification Framework} \label{sec:diver_identification_framework}
With our proposed diver identification algorithm, we develop a diver identification framework (see Fig.~\ref{fig:algoframe}) using robot operating system (ROS)~\cite{quigley2009ros}. It enables a robot to detect and identify divers, and it is designed to be compatible with both offline (models are trained before deployment) and online (models are trained during deployment) frameworks. 
The goal is to find a `target' diver where the target diver is either pre-assigned (offline framework) or assigned during mission (online framework).

We implement these frameworks using a state machine that utilizes a diver-relative position (DRP) estimator and a robot controller. These components are designed based on the Autonomous Diver-Relative Operator Configuration (ADROC) system introduced in~\cite{adroc}. Our state machine has the following eight states: INIT, SEARCH, APPROACH, DATA COLLECTION, ANGLE YAW, MODEL TRAINING, IDENTIFICATION, and CONCLUSION. 
The DRP estimator approximates the distance from the robot to the diver using bounding box detection (at a further distance) or pose estimation (at a closer distance) of the diver. When the distance information is available, \ie when the robot detects a diver, its motion is governed through a proportional-integral-derivative (PID) controller~\cite{nise2000pid}. 

\subsubsection{Offline Framework}
The state machine starts with the INIT state and immediately makes a transition to the SEARCH state until a distance information is available from the DRP estimator. Once available, the robot enters the APPROACH state. Now, the robot attempts to find and maintain a preset distance between itself and the diver. After reaching the desired distance, the robot enters the DATA COLLECTION state, where it computes the $45$-d ADR features from $F$ (a predefined number) filtered pose estimations. With the ($F\times 45$) ADR features, the robot goes into the IDENTIFICATION state since the models are already trained in this framework. If the target diver is found, the robot enters the CONCLUSION state. Otherwise, it enters the SEARCH state again after staying briefly in the ANGLE YAW state. We included the ANGLE YAW state to make the robot intelligently yaw away from the current diver to prevent performing identification on the same diver again.

\subsubsection{Online Framework}
It operates the same way as the offline framework up to the DATA COLLECTION state. Since the classification models are not pre-trained in this framework, the state machine enters the MODEL TRAINING state.
Here, the models (\eg SVM and KNN) are trained using ($F\mathcal{X}\times 45$) ADR features collected from $\mathcal{X}$ number of divers. 
Since we only train SVM and KNN classifiers in this state, the training time is acceptable on typical robotic hardware (\eg a mobile GPU or CPU). 
Once the training is done, a target diver will randomly be assigned. However, our framework has the flexibility to update the target diver as needed, \eg through specific hand gestures, or human-to-robot interaction. Then, the robot goes to the INIT state, and the state machine subsequently follows the same logic as the offline framework since the models are now trained.
\begin{table}[]
\centering
\caption{Comparison of average accuracy for the metric learning embedding network and the $10$ models on our UDI dataset. The values are in percentages.}
\begin{tabular}{lcccc}
\toprule
             & \multicolumn{2}{c}{\textbf{Embedding Network}}      & \multicolumn{2}{c}{\textbf{Models}}   \\
 \textbf{Dataset Type} & \textbf{All-Class}  & \textbf{Diver} & \textbf{All-Class}  & \textbf{Diver} \\
 \midrule
Training Accuracy & 97.78  & 97.89 & 97.21  & 97.97 \\
Testing Accuracy & 96.39  & 96.65 & 97.56  & 97.89 \\

\bottomrule
\end{tabular}
\label{tab:net_acc}
\vspace{-4mm}
\end{table}

\section{Experiments}\label{sec:exp}
This section describes experimental setups for evaluating our proposed algorithm including dataset creation, model building and training, and robot trials with divers. Human data collection and trials were conducted with the approval of the University of Minnesota’s Institutional Review Board (study reference no. 00013497).

\subsection{UDI Dataset}
To facilitate the learning of our proposed algorithm, we need to create a dataset that contains AD values of divers. Since it is extremely challenging to find and recruit many certified divers to collect such data in a closed-water environment (\eg pool), we collect the data from both swimmers and divers. In total, we have data from four swimmers and four divers. 
We ask the participants to take frontal-facing and standing-upright postures towards the camera, and we capture their full body images using GoPro cameras 
at varying distances. We collect a total of $16,557$ images across the eight participants. We then perform pose estimation on these images, filter out erroneous poses, and compute the ADR features. This gives us a total of $13,994$ ADR features and their corresponding labels, which we use to formulate the \textbf{underwater diver identification (UDI)} dataset. We employ a $80/20$ data split for training and testing. For the filtering step, as described in Sec.~\ref{sec:method}, we heuristically set $hw_{min}$, $k_{th}$, and $sw_{min}$ as $10$.

\subsection{Setup}
We use two versions of the UDI dataset to train our models: (1) \textbf{All-class} dataset (the entire UDI dataset) and (2) \textbf{Diver} dataset (a subset of the UDI dataset containing only divers). 
The first four models in Table~\ref{tab:all_models} do not include the embedding network in their structures and hence are trained directly with $45$-d raw ADR features. In contrast, the last six models include the pre-trained embedding network and therefore are trained with $16$-d features. 
We use scikit-learn libraries~\cite{pedregosa2011scikit} to implement the KNN and SVM models, and, PyTorch libraries~\cite{paszke2019pytorch} to implement the neural networks. For the KNN model, we set the neighbor size as $5$. 
Our metric learning embedding network is trained for $1,000$ epochs with a batch size of $512$, learning rate of $5\times 10^{-4}$, and margin, $m=0.3$. We use stochastic gradient descent (SGD) as the optimizer~\cite{ruder2016overview}. For the All\_NN and Diver\_NN models, the classification network is a two-layer neural network.

\begin{figure}  
    \vspace{2mm}
    \centering
    \scalebox{1.5}{\input{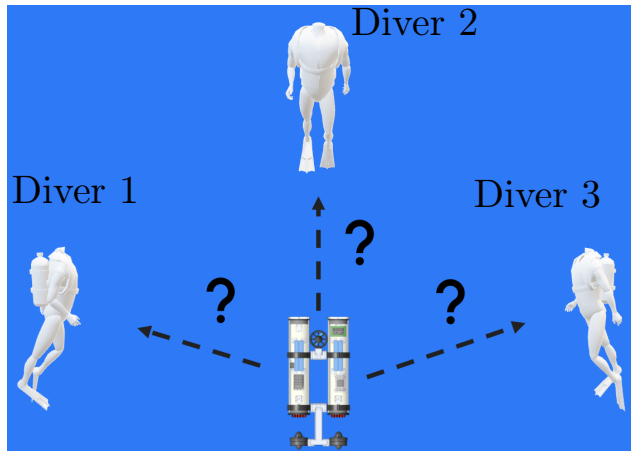}}
    \caption{Experiments setup: an AUV is positioned with three divers in a swimming pool where its goal is to identify a target diver.}  
    \label{fig:exp_setup}
    \vspace{-4mm}
\end{figure}

For offline framework, the embedding and classification networks are both trained before deployment. In contrast, the classification network is trained during mission in online framework although the embedding network is still pre-trained. Table~\ref{tab:net_acc} presents the average training and testing accuracy for both the embedding network and $10$ models presented in Table~\ref{tab:all_models}. Additionally, we train the models with features computed from either $50$ or $100$ frames to evaluate the correlation between the number of training data and the performance of each model.
To keep the online training time short, we do not consider training the Diver\_NN model.

\begin{table}[]
\centering
\vspace{2mm}
\caption{Results of $16$ robot trials conducted in closed-water environments for diver identification task where prediction accuracy = $\frac{(TP+TN)}{(TP+TN+FP+FN)}$.}
\begin{tabular}{c|cccc|cccc}
\toprule
\multirow{2}{*}{\textbf{\begin{tabular}[c]{@{}c@{}}Trial\\ No.\end{tabular}}} & \multicolumn{4}{c|}{\textbf{Our Method}} & \multicolumn{4}{c}{\textbf{Baseline}} \\
    & TP        & TN       & FP       & FN       & TP        & TN       & FP       & FN       \\ \midrule
1   &           &          & 1        &          &           &          &          & 1     \\
2   &   1        &          &         &          &  1         &          &          &      \\
3   &   1        &          &         &          &           &          &   1       & 1     \\
4   &   1        &   1       &         &          &           &   1       &          & 1     \\
5   &           &    1      &         &          &           &    1      &          &      \\
6   &           &    1      &         &          &  1         &          &    1      &      \\
7   &           &    1      &         &          &           &    1      &          & 1     \\
8   &           &          &         &    1      &           &          &     1     &      \\
9   &   1        &    2      &         &          &           &          &    2      & 1     \\
10   &           &    2      &         &          &           &          &    1      &      \\
11   &           &    1      &         &  1        &   1        &          &          & 1     \\
12   &  1         &          &         &          &           &          &          & 1     \\
13   &           &    1      &         &          &           &          &          & 1     \\
14   &  1         &    1      &        &          &    1       &          &    1      &      \\
15   &  1         &          &         &  1        &   1        &          &          & 1     \\
16   &           &          &        &    1      &      \multicolumn{4}{c}{no face detection}  \\
\midrule
Sum   &  7         &    11      &   1     &    4      &  5         &  3        &    7   & 9     \\
\midrule
Pred. Acc.   &  \multicolumn{4}{c|}{78.26\%}      &  \multicolumn{4}{c}{33.33\%}     \\
\bottomrule
\end{tabular}
\label{tab:result_summary}
\vspace{-4mm}
\end{table}

\subsection{Experimental Scenarios}
We use three divers in closed-water environments to perform our experiments using the offline and online frameworks. Fig.~\ref{fig:exp_setup} visualizes the setup for our experiments.
In the offline framework, our AUV collects ADR features from a diver and then feeds the features through pre-trained models and gets a prediction of diver identity. If the prediction is accurate, then the algorithm concludes and enables the robot to initiate an interaction with the diver. Otherwise, the robot turns to the next diver. The robot repeats this for all the divers present in the scene. We assume that the robot knows \textit{a priori} the total number of divers present in the scene.

In the online framework, the robot aims to identify a diver without using pre-trained classification models. To achieve this, first, the robot collects the ADR features from all the observable divers. It then trains six models online, to create highly separable clusters of divers. Once a target diver is assigned, the robot uses these newly trained models to identify the diver. Whenever there is a correct match, the algorithm concludes without checking the rest of the divers. If the robot is unable to find a correct match after checking all divers, it is considered a failure case. 

    \begin{figure*}
        \centering
        \vspace{2mm}
        \begin{subfigure}[t]{0.24\linewidth}
            \centering
            \includegraphics[width=\linewidth]{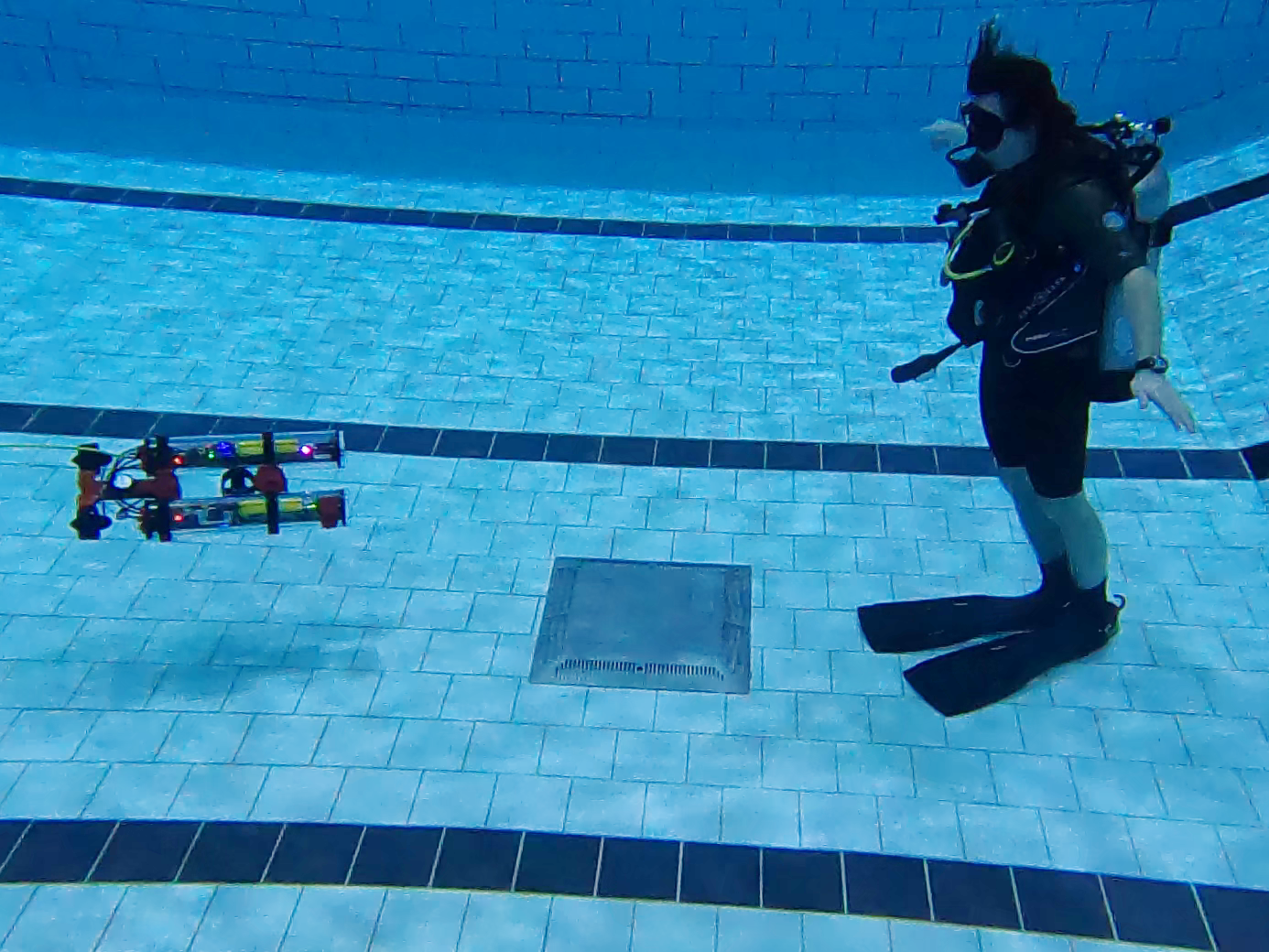}
            \caption{Third person view.}
            \label{fig:demo_statesa}
        \end{subfigure}
        \begin{subfigure}[t]{0.73\linewidth}
            \centering 
            \includegraphics[width=\linewidth]{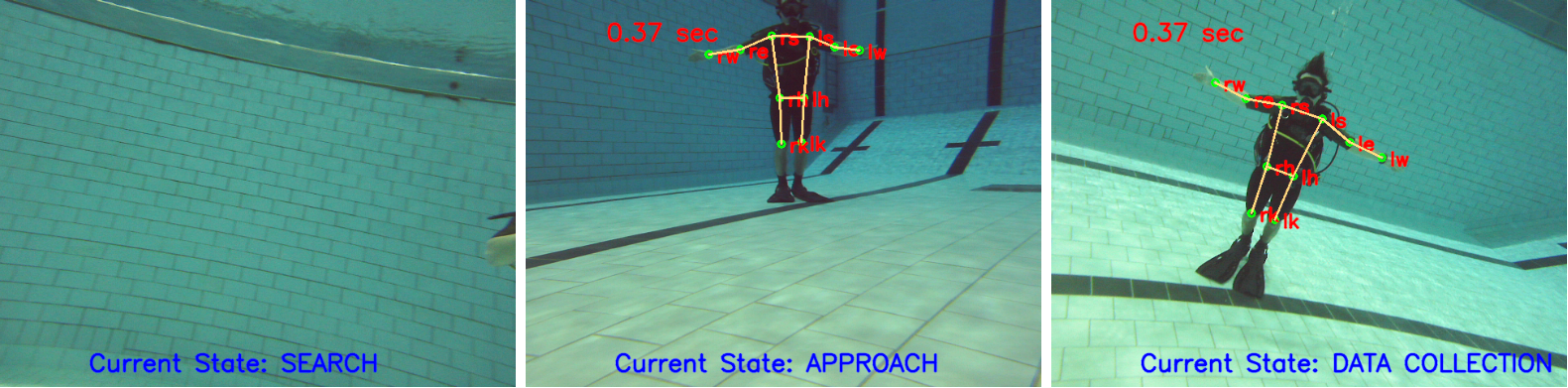}
            \caption{Robot's view during different states of the diver identification process.}
            \label{fig:demo_statesb}
        \end{subfigure}
                
        \caption{Demonstration of the proposed framework on a physical AUV platform in a closed-water environment. (\subref{fig:demo_statesa}) An external view of the diver identification process, (\subref{fig:demo_statesb}) some of the states in our framework from the robot's viewpoint.} 
        \label{fig:demo_states}
    \end{figure*}

    \begin{figure*}
        \centering
        \begin{subfigure}[b]{0.32\linewidth}
            \centering
            \includegraphics[width=\linewidth]{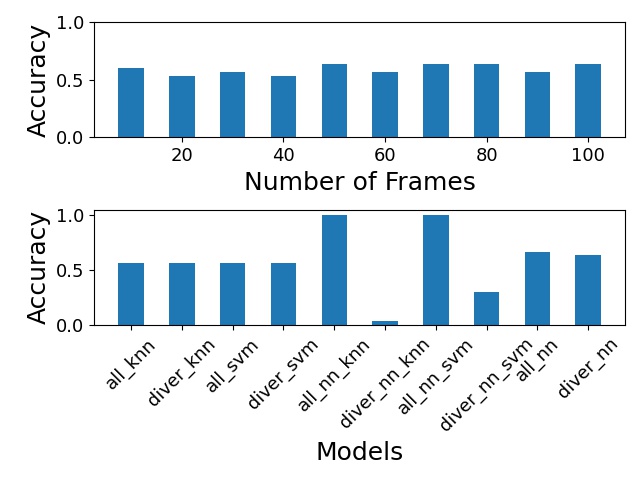}
            \caption{Trained Offline.}
            \label{fig:offlineexp}
        \end{subfigure}
        \hfill
        \begin{subfigure}[b]{0.32\linewidth}  
            \centering 
            \includegraphics[width=\linewidth]{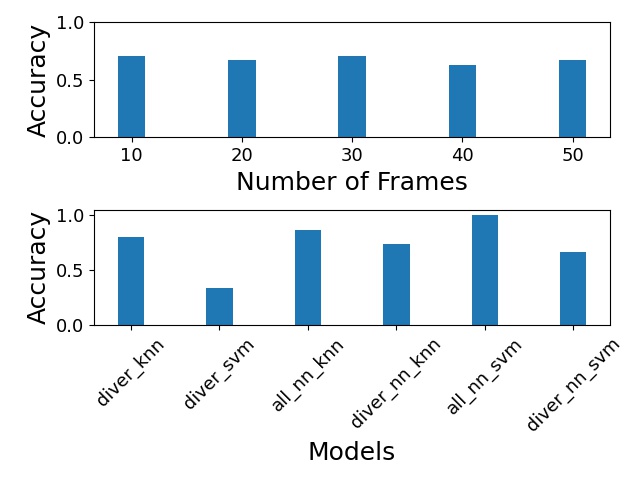}
            \caption{Trained Online with $50$ data points.}
            \label{fig:online50exp}
        \end{subfigure}
        \hfill
        \begin{subfigure}[b]{0.32\linewidth}   
            \centering 
            \includegraphics[width=\linewidth]{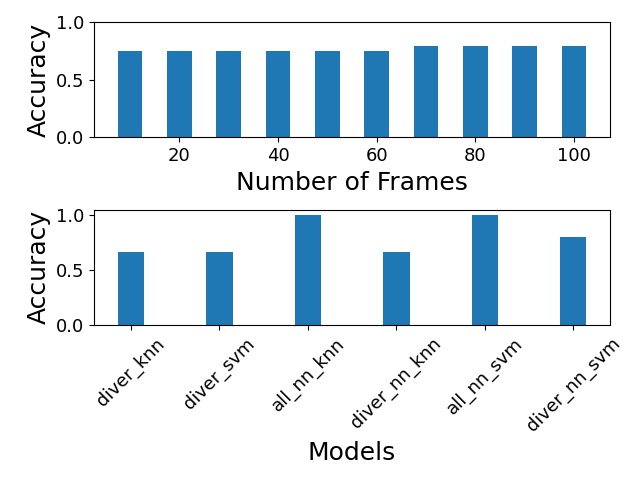}
            \caption{Trained Online with $100$ data points.}
            \label{fig:online100exp}
        \end{subfigure}
        \caption{Average accuracy from our experiments: (Top row): average accuracy for a specific number of test frames across all models, (Bottom row): average accuracy of each model across all frames.} 
        \label{fig:model_accexp}
        \vspace{-4mm}
    \end{figure*}

\section{Results}

We deploy the system onboard the LoCO AUV~\cite{edge2020design} in closed-water environments to conduct $16$ robot trials. During each of these trials, the AUV starts by searching for a diver, approaching them for data collection, and then performing identification. The AUV performs these actions completely autonomously using our proposed diver identification framework. To our knowledge, the only other work that attempts to address the problem of diver identification is presented in~\cite{hong2020visual} where the authors propose to use facial cues to identify different divers. We consider this work to be the baseline method for evaluating the performance of our proposed system. During the trials, we compute \textit{true positive, TP}, \textit{true negative, TN}, \textit{false positive, FP}, and \textit{false negative, FN}. 
Table~\ref{tab:result_summary} summarizes the results of these trials. From the table, we notice that our proposed method accurately identifies divers in $18$ out of $23$ instances, yielding an average prediction accuracy of $78.26\%$. In comparison, the baseline method achieves only a $33.33\%$ average prediction accuracy and fails to detect any diver faces during robot trial no. $16$. 
This demonstrates that our proposed method not only reliably identifies divers but also outperforms the only other state-of-the-art method in the diver identification task. Fig.~\ref{fig:demo_states} displays several snapshots of the proposed diver identification method taken during the robot trials. In the following sections, we will discuss the performance of the proposed system in both offline and online frameworks.

\subsection{Performance of the Offline Framework}
We test the performance of all $10$ models pre-trained with the \textbf{All-class} and \textbf{Diver} datasets. Fig.~\ref{fig:offlineexp} presents the quantitative results of our experiments. 
The bottom plot in Fig.~\ref{fig:offlineexp} shows the average accuracy of each model across all numbers of test frames. From the figure, the superior performance of All\_NN\_KNN and All\_NN\_SVM is evident. Finally, we calculate the average accuracy across all the models (the top plot in Fig.~\ref{fig:offlineexp}) and notice that the accuracy remains almost the same. This suggests that All\_NN\_KNN and All\_NN\_SVM can even make faster predictions using a lower number of frames while maintaining their high accuracy. 

\subsection{Performance of the Online Framework}
We first use $50$ frames \textit{per} diver to train the models and evaluate the diver identification performance. From Fig.~\ref{fig:online50exp} (bottom), we notice that All\_NN\_SVM performs the best and All\_NN\_KNN shows comparable performance. That means the metric learning technique is indeed benefiting the identification task. Even if we change the number of frames \textit{per} diver to train the models to $100$, we see similar trends. From Fig.~\ref{fig:online100exp} (bottom), we see that both the All\_NN\_SVM and All\_NN\_KNN achieve the highest accuracy. Finally, from Fig.~\ref{fig:online50exp} (top) and \ref{fig:online100exp} (top), we notice that the number of frames used for making inferences does not have any significant effect on the system's performance for the online framework, as was the case for the offline framework. 

As these results suggest, our diver identification framework can successfully identify divers with both the offline and online frameworks. In both cases, the All\_NN\_KNN and All\_NN\_SVM models outperform the other models. Especially, the All\_NN\_SVM model shows consistently accurate identification performance regardless of the type of the framework and the number of frames for training and inference. Additionally, we see that the models with our embedding network achieve higher identification accuracy than the models without the network. More importantly, our results demonstrate that the ADR features are distance and photometric-invariant considering the fact that (1) our data has been captured at varying distances, and (2) images from the camera ($0.9$-megapixel, $720$p resolution) that we used to build our training dataset with and the robot camera ($2$-megapixel, $1080$p), are significantly different. Fig.~\ref{fig:demo_statesa} and Fig.~\ref{fig:demo_statesb} show sample pictures taken by these two cameras, respectively.
\section{Conclusions}
In this work, we present the design and implementation of a diver identification algorithm for AUVs using anthropometric data ratios (ADRs) as features, which are invariant to distance and photometric conditions. 
With experiments performed during closed-water trials, we show that our proposed algorithm enables an AUV to extract the ADR features from divers using a pose estimation technique and use them to identify a target diver. We incorporate our diver identification algorithm within both offline and online frameworks. This enables the AUV to identify divers even if we do not have their data in our dataset prior to deployments. The key advantages of our method are: it only requires a monocular camera to identify a diver, and it can be deployed on-board a physical AUV. We believe our proposed method will enable secure human-robot collaborative missions by successfully identifying divers. 
We are extending our algorithm to identify multiple divers simultaneously and planning to use our framework in collaborative missions for aquatic invasive species mapping and marine debris removal.

\bibliographystyle{unsrt}  
\bibliography{refer}

\begin{thebibliography}{10}

\bibitem{bayat2017environmental}
Behzad Bayat, Naveena Crasta, Alessandro Crespi, Ant{\'o}nio~M Pascoal, and Auke Ijspeert.
\newblock {Environmental Monitoring using Autonomous Vehicles: A Survey of Recent Searching Techniques}.
\newblock {\em Current Opinion in Biotechnology}, 45:76--84, 2017.

\bibitem{islam2019diverfollowing}
Md.~Jahidul Islam, Michael Fulton, and Junaed Sattar.
\newblock {Toward a Generic Diver-Following Algorithm: Balancing Robustness and Efficiency in Deep Visual Detection}.
\newblock {\em IEEE Robotics and Automation Letters}, 4(1):113--120, 2019.

\bibitem{bingham2010robotic}
Brian Bingham, Brendan Foley, Hanumant Singh, Richard Camilli, Katerina Delaporta, Ryan Eustice, et~al.
\newblock {Robotic Tools for Deep Water Archaeology: Surveying an Ancient Shipwreck with an Autonomous Underwater Vehicle}.
\newblock {\em {Journal of Field Robotics (JFR)}}, 27(6):702--717, 2010.

\bibitem{sattar2007fourier}
Junaed Sattar, Eric Bourque, Philippe Giguere, and Gregory Dudek.
\newblock {Fourier Tags: Smoothly Degradable Fiducial Markers for Use in Human-Robot Interaction}.
\newblock In {\em Fourth Canadian Conference on Computer and Robot Vision (CRV '07)}, pages 165--174, 2007.

\bibitem{islam2018dynamic}
Md~Jahidul Islam, Marc Ho, and Junaed Sattar.
\newblock {Dynamic Reconfiguration of Mission Parameters in Underwater Human-Robot Collaboration}.
\newblock In {\em 2018 IEEE International Conference on Robotics and Automation (ICRA)}, pages 6212--6219. IEEE, 2018.

\bibitem{mivskovic2016human}
Nikola Mi{\v{s}}kovi{\'c}, Murat Egi, Dula Nad, Antonio Pascoal, Luis Sebastiao, and Marco Bibuli.
\newblock {Human-Robot Interaction Underwater: Communication and Safety Requirements}.
\newblock In {\em 2016 IEEE Third Underwater Communications and Networking Conference (UComms)}, pages 1--5. IEEE, 2016.

\bibitem{hong2020visual}
Jungseok Hong, Sadman~Sakib Enan, Christopher Morse, and Junaed Sattar.
\newblock {Visual Diver Face Recognition for Underwater Human-Robot Interaction}.
\newblock {\em arXiv preprint arXiv:2011.09556}, 2020.

\bibitem{bertillon1896signaletic}
Alphonse Bertillon.
\newblock {Signaletic Instructions, including the Theory and Practice of Anthropometrical Identification}.
\newblock {\em Nature}, pages 569--570, October 1896.

\bibitem{drozdowski2019computational}
Pawel Drozdowski, Christian Rathgeb, and Christoph Busch.
\newblock {Computational Workload in Biometric Identification Systems: An Overview}.
\newblock {\em IET Biometrics}, 8(6):351--368, 2019.

\bibitem{wilaiprasitporn2020affective}
Theerawit Wilaiprasitporn, Apiwat Ditthapron, Karis Matchaparn, Tanaboon Tongbuasirilai, Nannapas Banluesombatkul, and Ekapol Chuangsuwanich.
\newblock {Affective EEG-Based Person Identification Using the Deep Learning Approach}.
\newblock {\em IEEE Transactions on Cognitive and Developmental Systems}, 12(3):486--496, 2020.

\bibitem{kortli2020face}
Yassin Kortli, Maher Jridi, Ayman Al~Falou, and Mohamed Atri.
\newblock {Face Recognition Systems: A Survey}.
\newblock {\em Sensors}, 20(2):342, 2020.

\bibitem{deng2019arcface}
J.~{Deng}, J.~{Guo}, N.~{Xue}, and S.~{Zafeiriou}.
\newblock {ArcFace: Additive Angular Margin Loss for Deep Face Recognition}.
\newblock In {\em 2019 IEEE/CVF Conference on Computer Vision and Pattern Recognition (CVPR)}, pages 4685--4694, 2019.

\bibitem{yang2016wider}
Shuo Yang, Ping Luo, Chen-Change Loy, and Xiaoou Tang.
\newblock {Wider Face: A Face Detection Benchmark}.
\newblock In {\em Proceedings of the IEEE conference on computer vision and pattern recognition}, pages 5525--5533, 2016.

\bibitem{guo2016ms}
Yandong Guo, Lei Zhang, Yuxiao Hu, Xiaodong He, and Jianfeng Gao.
\newblock {MS-Celeb-1M: A Dataset and Benchmark for Large-Scale Face Recognition}.
\newblock In {\em Computer Vision--ECCV 2016: 14th European Conference, Amsterdam, The Netherlands, October 11-14, 2016, Proceedings, Part III 14}, pages 87--102. Springer, 2016.

\bibitem{bae2023digiface}
Gwangbin Bae, Martin de~La~Gorce, Tadas Baltru{\v{s}}aitis, Charlie Hewitt, Dong Chen, Julien Valentin, Roberto Cipolla, and Jingjing Shen.
\newblock {DigiFace-1M: 1 Million Digital Face Images for Face Recognition}.
\newblock In {\em Proceedings of the IEEE/CVF Winter Conference on Applications of Computer Vision}, pages 3526--3535, 2023.

\bibitem{bengio2017deep}
Yoshua Bengio, Ian Goodfellow, and Aaron Courville.
\newblock {\em {Deep learning}}, volume~1.
\newblock MIT press Cambridge, MA, USA, 2017.

\bibitem{khan2021gait}
Muhammad~Hassan Khan, Muhammad~Shahid Farid, and Marcin Grzegorzek.
\newblock {Vision-based Approaches Towards Person Identification using Gait}.
\newblock {\em Computer Science Review}, 42:100432, 2021.

\bibitem{xia2019visual}
Youya Xia and Junaed Sattar.
\newblock {Visual Diver Recognition for Underwater Human-Robot Collaboration}.
\newblock In {\em 2019 International Conference on Robotics and Automation (ICRA)}, pages 6839--6845. IEEE, 2019.

\bibitem{de2020realtime}
Karin De~Langis and Junaed Sattar.
\newblock {Real-Time Multi-Diver Tracking and Re-identification for Underwater Human-Robot Collaboration}.
\newblock In {\em 2020 IEEE International Conference on Robotics and Automation (ICRA)}, pages 11140--11146. IEEE, 2020.

\bibitem{munsell2012person}
Brent~C Munsell, Andrew Temlyakov, Chengzheng Qu, and Song Wang.
\newblock {Person Identification Using Full-Body Motion and Anthropometric Biometrics From Kinect Videos}.
\newblock In {\em Computer Vision--ECCV 2012. Workshops and Demonstrations: Florence, Italy, October 7-13, 2012, Proceedings, Part III 12}, pages 91--100. Springer, 2012.

\bibitem{andersson2015person}
Virginia Andersson and Ricardo Araujo.
\newblock {Person Identification Using Anthropometric and Gait Data from Kinect Sensor}.
\newblock {\em Proceedings of the AAAI Conference on Artificial Intelligence}, 29(1), Feb. 2015.

\bibitem{openpose}
Zhe Cao, Tomas Simon, Shih-En Wei, and Yaser Sheikh.
\newblock {Realtime Multi-Person 2D Pose Estimation Using Part Affinity Fields}.
\newblock In {\em Proceedings of the IEEE Conference on Computer Vision and Pattern Recognition (CVPR)}, July 2017.

\bibitem{trtpose}
NVIDIA~AI IOT.
\newblock trt\_pose.
\newblock \url{https://github.com/NVIDIA-AI-IOT/trt_pose}, 2019.

\bibitem{lugaresi2019mediapipe}
Camillo Lugaresi, Jiuqiang Tang, Hadon Nash, Chris McClanahan, Esha Uboweja, Michael Hays, Fan Zhang, Chuo-Ling Chang, Ming~Guang Yong, Juhyun Lee, et~al.
\newblock {MediaPipe: A Framework for Building Perception Pipelines}.
\newblock {\em arXiv preprint arXiv:1906.08172}, 2019.

\bibitem{mathis2018deeplabcut}
Alexander Mathis, Pranav Mamidanna, Kevin~M Cury, Taiga Abe, Venkatesh~N Murthy, Mackenzie~Weygandt Mathis, and Matthias Bethge.
\newblock {DeepLabCut: Markerless Pose Estimation of User-defined Body Parts with Deep Learning}.
\newblock {\em Nature Neuroscience}, 21(9):1281--1289, 2018.

\bibitem{wang2020deep}
Jingdong Wang, Ke~Sun, Tianheng Cheng, Borui Jiang, Chaorui Deng, Yang Zhao, Dong Liu, Yadong Mu, Mingkui Tan, Xinggang Wang, et~al.
\newblock {Deep High-Resolution Representation Learning for Visual Recognition}.
\newblock {\em IEEE transactions on pattern analysis and machine intelligence}, 43(10):3349--3364, 2020.

\bibitem{gordon1989anthropometric}
Claire~C Gordon, Thomas Churchill, Charles~E Clauser, Bruce Bradtmiller, John~T McConville, Ilse Tebbetts, and Robert~A Walker.
\newblock {Anthropometric Survey of US Army Personnel: Summary Statistics, Interim Report for 1988}.
\newblock Technical report, Anthropology Research Project Inc Yellow Springs OH, 1989.

\bibitem{yang2006distance}
Liu Yang and Rong Jin.
\newblock {Distance Metric Learning: A Comprehensive Survey}.
\newblock {\em Michigan State University}, 2(2):4, 2006.

\bibitem{peterson2009k}
Leif~E Peterson.
\newblock {K-Nearest Neighbor}.
\newblock {\em Scholarpedia}, 4(2):1883, 2009.

\bibitem{noble2006support}
William~S Noble.
\newblock {What Is a Support Vector Machine?}
\newblock {\em Nature Biotechnology}, 24(12):1565--1567, 2006.

\bibitem{schroff2015facenet}
Florian Schroff, Dmitry Kalenichenko, and James Philbin.
\newblock {FaceNet: A Unified Embedding for Face Recognition and Clustering}.
\newblock In {\em Proceedings of the IEEE conference on computer vision and pattern recognition}, pages 815--823, 2015.

\bibitem{van2008visualizing}
Laurens Van~der Maaten and Geoffrey Hinton.
\newblock {Visualizing Data using t-SNE.}
\newblock {\em Journal of machine learning research}, 9(11), 2008.

\bibitem{quigley2009ros}
Morgan Quigley, Ken Conley, Brian Gerkey, Josh Faust, Tully Foote, Jeremy Leibs, Rob Wheeler, Andrew~Y Ng, et~al.
\newblock {ROS: An Open-Source Robot Operating System}.
\newblock In {\em ICRA workshop on open source software}, volume~3, page~5. Kobe, Japan, 2009.

\bibitem{adroc}
Michael Fulton, Jungseok Hong, and Junaed Sattar.
\newblock {Using Monocular Vision and Human Body Priors for AUVs to Autonomously Approach Divers}.
\newblock In {\em 2022 International Conference on Robotics and Automation (ICRA)}, pages 1076--1082, 2022.

\bibitem{nise2000pid}
Norman~S Nise.
\newblock {\em Control systems engineering}.
\newblock John Wiley \& Sons, 2020.

\bibitem{pedregosa2011scikit}
Fabian Pedregosa, Ga{\"e}l Varoquaux, Alexandre Gramfort, Vincent Michel, Bertrand Thirion, Olivier Grisel, Mathieu Blondel, Peter Prettenhofer, Ron Weiss, Vincent Dubourg, et~al.
\newblock {Scikit-learn: Machine Learning in Python}.
\newblock {\em The Journal of Machine Learning Research}, 12:2825--2830, 2011.

\bibitem{paszke2019pytorch}
Adam Paszke, Sam Gross, Francisco Massa, Adam Lerer, James Bradbury, Gregory Chanan, Trevor Killeen, Zeming Lin, Natalia Gimelshein, Luca Antiga, Alban Desmaison, Andreas Kopf, Edward Yang, Zachary DeVito, Martin Raison, Alykhan Tejani, Sasank Chilamkurthy, Benoit Steiner, Lu~Fang, Junjie Bai, and Soumith Chintala.
\newblock {PyTorch: An Imperative Style, High-Performance Deep Learning Library}.
\newblock In {\em Advances in Neural Information Processing Systems}, pages 8024--8035. Curran Associates, Inc., 2019.

\bibitem{ruder2016overview}
Sebastian Ruder.
\newblock {An overview of gradient descent optimization algorithms}.
\newblock {\em arXiv preprint arXiv:1609.04747}, 2016.

\bibitem{edge2020design}
Chelsey {Edge}, Sadman~Sakib {Enan}, Michael {Fulton}, Jungseok {Hong}, Jiawei {Mo}, Kimberly {Barthelemy}, Hunter {Bashaw}, Berik {Kallevig}, Corey {Knutson}, Kevin {Orpen}, and Junaed {Sattar}.
\newblock {Design and Experiments with LoCO AUV: A Low Cost Open-Source Autonomous Underwater Vehicle*}.
\newblock In {\em 2020 IEEE/RSJ International Conference on Intelligent Robots and Systems (IROS)}, pages 1761--1768, 2020.

\end{thebibliography}

\end{document}